\documentclass{article}
\usepackage[final,dblblindworkshop]{neurips_2026}
\setcitestyle{round}
\usepackage[utf8]{inputenc}
\usepackage[T1]{fontenc}
\usepackage{hyperref}
\usepackage{url}
\usepackage{booktabs}
\usepackage{amsfonts}
\usepackage{amsmath}
\usepackage{amssymb}
\usepackage{nicefrac}
\usepackage{microtype}
\usepackage{xcolor}
\usepackage{graphicx}
\usepackage{subcaption}
\usepackage{algorithm}
\usepackage{algorithmic}
\usepackage{multirow}
\usepackage{wrapfig}

\title{Random Initialization as a Baseline for Time Series Foundation Models}

\author{
  Pinar Sungu Isiacik, Georgiana Ifrim\\
  University College Dublin\\
  Dublin, Ireland\\
  \texttt{pinar.sunguisiacik@ucdconnect.ie, georgiana.ifrim@ucd.ie}
}

\begin{document}

\maketitle

\begin{abstract}
Time series foundation models rely on self-supervised objectives, such as masked reconstruction or contrastive learning, to build reusable representations for downstream tasks such as multivariate time series classification (MTSC). In this work, we test how much pretrained weights improve over a baseline that requires no pretraining: the same architecture with randomly initialized weights. We evaluate seven publicly released instances from four time series foundation model families on the UEA MTSC archive\footnote{\url{https://www.timeseriesclassification.com/}}, with per-model exclusions where evaluation cannot be completed due to computational constraints. For each instance, we compare the pretrained weights against five randomly initialized variants using four downstream classifiers on frozen
representations: two linear classifiers (ridge classifier and logistic regression) and two nonlinear classifiers (RBF SVM and random forest), yielding 183 paired dataset-level comparisons. Pretrained variants achieve higher mean accuracy across all seven instances, but the mean advantage is modest, ranging from 1.43 to 5.19 percentage points, with an overall average gain of 2.58 percentage
points. Under per-instance Wilcoxon signed-rank tests, three of seven raw $p$-values fall below 0.05, but after Holm correction for multiple comparisons, only NuTime remains statistically significant. A secondary pooled analysis across all 183 pairs yields $p < 0.001$, while the corrected per-instance tests remain the primary statistical evidence. We further compare against three established time-series-specific classifiers: ROCKET, Hydra, and Quant, which remain competitive with the pretrained foundation models without requiring large-scale pretraining. Together, these results motivate a more careful assessment of the computational
cost of large-scale pretraining relative to its downstream benefit for MTSC and support the systematic inclusion of architecture-matched random initialization and non-pretrained baselines in future evaluations.  Code, experiment configurations, and complete results are available in a repository.\footnote{\url{https://github.com/mlgig/tsfm-random-init-baselines.git}}
\end{abstract}

\section{Introduction}
\label{sec:intro}

Labeled time series can be costly and time-consuming to obtain, particularly when annotation requires domain expertise, motivating growing interest in self-supervised learning to reduce dependence on labeled data~\citep{zhang2024ssl}. Time series foundation models (TSFMs) apply this paradigm at scale by pretraining encoders on large collections of unlabeled sequences for transfer to downstream tasks. The models considered in this study span multiple architectures and model scales.

We focus on multivariate time series classification (MTSC), a comparatively less studied setting than univariate classification~\citep{ruiz2021bakeoff}. MTSC is particularly interesting for TSFMs because observations contain multiple channels, while many current models are pretrained on univariate sequences and process multivariate inputs independently per channel~\citep{pinar2025fm4mtsc}. Their frozen representations are then typically aggregated before downstream classification. This raises a basic question: how much of the resulting performance comes from pretrained weights rather than the architecture and aggregation pipeline?

This distinction matters because randomly initialized networks can themselves produce useful high-dimensional representations~\citep{rahimi2007}. ROCKET~\citep{rocket}, for example, demonstrates strong time series classification using random convolutional features. If an architecture-matched randomly initialized TSFM approaches the performance of its pretrained counterpart, then part of the downstream performance may be attributable to the architecture and aggregation pipeline rather than the pretrained weights alone. We therefore evaluate both this controlled comparison and whether the resulting performance is competitive with established time-series-specific classifiers.

Related evidence from computer vision shows that models trained from random initialization can sometimes approach counterparts initialized from pretrained weights~\citep{he2019rethinking,raghu2019transfusion}. More directly, \citet{yang2026} report comparable performance for pretrained and randomly initialized Mantis representations~\citep{mantis} on four EEG classification benchmarks. However, it remains unclear whether this pattern extends across multiple TSFM families and model scales, different downstream classifier capacities, and the broader range of general-purpose MTSC datasets considered here.

We evaluate seven instances from four TSFM families: MOMENT~\citep{moment}, Mantis~\citep{mantis}, UniTS~\citep{units}, and NuTime~\citep{nutime}, on the UEA MTSC archive~\citep{uea}, together with ROCKET~\citep{rocket}, Hydra~\citep{hydra}, and Quant~\citep{quant}. Our contributions are:
\begin{itemize}
    \item A controlled comparison of pretrained and architecture-matched random TSFMs across seven model instances, four downstream classifiers spanning linear and nonlinear decision functions, and five random seeds.
    
    \item Evidence that pretraining yields a modest mean improvement of 1.43-5.19 percentage points per model instance, with an overall average gain of 2.58 percentage points across 183 paired comparisons. After Holm correction for multiple comparisons, only NuTime retains a statistically significant per-instance advantage.
    
    \item Evidence that the pretrained and random difference remains modest on average across both linear and nonlinear downstream classifiers, while established non-pretrained methods remain competitive. These results motivate the inclusion of both architecture-matched random initialization and strong time-series-specific methods as controls when evaluating large-scale time series pretraining.
\end{itemize}
\section{Related Work}
\label{sec:related}

Self-supervised time series representation learning commonly relies on masked reconstruction or contrastive objectives~\citep{zerveas2021,ts2vec,tfc}. These approaches underpin several time series foundation models considered in this work. MOMENT~\citep{moment} uses masked reconstruction, Mantis~\citep{mantis} and MantisV2~\citep{mantisv2} use contrastive pretraining, NuTime~\citep{nutime} adopts a BYOL-based self-distillation objective, and UniTS~\citep{units} combines multiple tasks within a unified backbone. Other foundation models primarily target forecasting~\citep{chronos,timesfm,moirai}, whereas we focus on multivariate time series classification.

Recent work has begun to examine the benefits of large-scale time series pretraining. \citet{pinar2025fm4mtsc} benchmark pretrained foundation models for MTSC and show that many process multivariate inputs in a channel-independent manner, but do not compare against architecture-matched random initialization. Related comparisons include \citet{yang2026} for EEG classification and TSFM-Bench~\citep{tsfmbench} for few-shot forecasting. We instead study frozen MTSC representations across multiple TSFM families and model scales. Related forecasting studies also question whether model complexity and pretraining consistently improve performance~\citep{zeng2023,tan2024}.

Our study also relates to random-feature and efficient time-series classification methods. Random feature mappings can provide useful representations without learned feature parameters~\citep{rahimi2007}, while randomly initialized neural networks induce non-trivial kernel structures~\citep{jacot2018}. ROCKET~\citep{rocket} demonstrates this principle for time series using random convolutional kernels, with related approaches such as MiniRocket and MultiRocket further improving efficiency and performance~\citep{minirocket,multirocket}. We additionally consider Hydra~\citep{hydra}, which uses competing convolutional kernels, and Quant~\citep{quant}, a lightweight interval-based method using quantile features. These methods provide complementary non-pretrained reference points, while our architecture-matched random initialization experiments isolate the contribution of pretrained weights by holding the model architecture and representation pipeline fixed.
\section{Experiments}
\label{sec:experiments}

Our experiments are designed around three questions. First, how much downstream performance is attributable specifically to pretrained weights when the architecture and representation pipeline are held fixed? Second, does this conclusion depend on the downstream classifier? Third, how do pretrained foundation-model representations compare with established methods designed specifically for time series classification?

\subsection{Foundation models}

We evaluate seven publicly released instances from four time series foundation model families: MOMENT~\citep{moment} (Small, Base, and Large), Mantis~\citep{mantis}, MantisV2~\citep{mantisv2}, UniTS~\citep{units}, and NuTime~\citep{nutime}. For each instance, we use the released pretrained checkpoint and its corresponding official implementation. Complete per-dataset and classifier-specific results for all evaluated instances are reported in Appendix~\ref{sec:complete_results}.

\subsection{Architecture-matched random initialization}

To isolate the contribution of pretraining from that of the architecture, we construct five randomly initialized variants of each model instance using seeds $\{0, 7, 42, 123, 2026\}$. Each variant uses the same model configuration as the corresponding pretrained model and is initialized using the default procedure of the official implementation. No pretrained encoder parameters are retained in the random condition, and the randomly initialized models are not pretrained or fine-tuned.

Pretrained and randomly initialized variants use the same preprocessing, multivariate inference, representation extraction, aggregation, and downstream evaluation procedures. The encoder remains frozen in both conditions, and the encoder weights are the experimental variable.

\subsection{Downstream representation evaluation}

We evaluate on the 30-dataset UEA MTSC archive~\citep{uea} using the standard training and test splits. A model-dataset result is included in the aggregate analysis only when all four downstream classifier results are available under both the pretrained and randomly initialized conditions. This yields 26 paired datasets for each MOMENT instance, Mantis, MantisV2, and NuTime, and 27 for UniTS. Each per-instance comparison is restricted to its available paired datasets, while the cross-method comparison uses the 24 datasets shared by all methods. Complete dataset coverage is reported in Appendix~\ref{app:dataset_coverage}.

For each model, condition, and dataset, we extract frozen representations for the training and test splits. We evaluate these representations using two linear classifiers, ridge classification and logistic regression, and two nonlinear classifiers, RBF SVM and random forest. This allows us to examine whether the pretrained-random difference depends on the downstream classifier.

For all four classifiers, hyperparameters are selected using the same three-fold stratified cross-validation procedure on the UEA training split. The folds are shuffled using \texttt{random\_state=42}. Standardization is included within the cross-validation pipeline for ridge classification, logistic regression, and RBF SVM. After hyperparameter selection, each classifier is refitted on the complete training split and evaluated once on the standard UEA test split. The hyperparameter grids are reported in Appendix~\ref{sec:implementation}.

\subsection{Time-series-specific reference methods}

Architecture-matched random initialization measures the contribution of pretrained weights within a fixed foundation-model architecture, but it does not establish how the resulting representations compare with methods designed directly for time series classification. We therefore additionally evaluate ROCKET~\citep{rocket}, Hydra~\citep{hydra}, and Quant~\citep{quant} on the UEA archive.

We implement these methods using the \texttt{aeon} library~\citep{middlehurst2024aeon}, with their default configurations and without dataset-specific tuning. Each method is fitted on the standard UEA training split and evaluated on the corresponding test split.

For direct comparison, we report results on the common subset of datasets for which all seven pretrained foundation-model instances and all three time-series-specific methods have valid results. This allows mean accuracy, average rank, and dataset wins to be compared without differences in dataset coverage. Complete results are provided in Appendix~\ref{sec:ts_results}.

\subsection{Statistical analysis}

For each model instance, dataset, and classifier, random-initialization accuracy is averaged across the five encoder initialization seeds. We then average across the four downstream classifiers to obtain one paired pretrained and random-initialization score per dataset.

For each model instance, we compare the paired dataset-level scores using a two-sided Wilcoxon signed-rank test~\citep{demsar2006}. We apply Holm's step-down correction to the seven per-instance tests at $\alpha=0.05$ and report both raw and adjusted $p$-values. To describe the magnitude and consistency of the difference, we additionally report the mean and median paired accuracy difference and the number of pretraining wins, ties, and losses across datasets. A win indicates that the pretrained model achieves higher accuracy than its randomly initialized counterpart, while a loss indicates the opposite. These statistics are computed from the same paired, unrounded accuracies used in the Wilcoxon tests. We additionally report unadjusted 95\% percentile bootstrap confidence intervals for the mean paired difference using 10,000 paired dataset-level resamples; the Holm-corrected Wilcoxon tests remain our primary inferential analysis.

As a secondary analysis, we report a pooled Wilcoxon signed-rank test across all model-dataset pairs. Because datasets recur across model instances, this pooled test is interpreted as complementary to the per-instance analyses. The comparison with ROCKET, Hydra, and Quant is reported descriptively using mean accuracy, average rank, and dataset wins.
\section{Results and Discussion}
\label{sec:results}

\subsection{How much does pretraining contribute?}

Table~\ref{tab:architecture_matched} shows that pretraining improves mean accuracy for all seven model instances, with gains of 1.43-5.19 percentage points and an overall mean gain of 2.58 points across 183 paired comparisons. Median gains range from 0.40 to 4.20 points; overall, pretraining records 127 wins, 3 ties, and 53 losses.

\begin{table}[htbp]
\centering
\caption{Architecture-matched comparison of pretrained and randomly initialized foundation models. Accuracy is averaged across four downstream classifiers. Random-init values are mean $\pm$ standard deviation across five encoder seeds. Mean $\Delta$ includes an unadjusted 95\% paired-bootstrap confidence interval; W/T/L denotes pretrained wins/ties/losses.}
\label{tab:architecture_matched}
\vspace{0.5em}
\scriptsize
\setlength{\tabcolsep}{4.0pt}
\begin{tabular}{l c c c c c c c c}
\toprule
\textbf{Instance} & $n$ & \textbf{Pretrained} & \textbf{Random init} & \textbf{Mean $\Delta$ [95\% CI]} & \textbf{Median $\Delta$} & \textbf{W/T/L} & $p_{\text{raw}}$ & $p_{\text{Holm}}$ \\
\midrule
MOMENT-Small & 26 & 0.703 & $0.680 \pm 0.002$ & $+2.29\,[-1.20,\,5.46]$ & +2.40 & 21/1/4 & 0.011 & 0.066 \\
MOMENT-Base & 26 & 0.698 & $0.684 \pm 0.003$ & $+1.43\,[-2.02,\,4.67]$ & +1.20 & 17/0/9 & 0.143 & 0.233 \\
MOMENT-Large & 26 & 0.702 & $0.684 \pm 0.005$ & $+1.76\,[-1.73,\,4.82]$ & +2.35 & 16/0/10 & 0.075 & 0.225 \\
Mantis & 26 & 0.728 & $0.702 \pm 0.002$ & $+2.56\,[-0.29,\,5.45]$ & +2.60 & 18/1/7 & 0.053 & 0.210 \\
MantisV2 & 26 & 0.721 & $0.695 \pm 0.003$ & $+2.58\,[0.16,\,5.08]$ & +0.80 & 16/0/10 & 0.116 & 0.233 \\
UniTS & 27 & 0.651 & $0.628 \pm 0.003$ & $+2.24\,[0.80,\,3.83]$ & +0.40 & 18/1/8 & 0.021 & 0.107 \\
NuTime & 26 & 0.626 & $0.574 \pm 0.005$ & $+5.19\,[2.80,\,7.85]$ & +4.20 & 21/0/5 & $<0.001$ & 0.002 \\
\midrule
Pooled & 183 & 0.689 & 0.664 & +2.58 & +2.00 & 127/3/53 & $<0.001$ & -- \\
\bottomrule
\end{tabular}
\end{table}

Three raw per-instance tests have $p<0.05$, but only NuTime remains significant after Holm correction. The pooled test yields $p<0.001$; since datasets recur across model instances, the corrected per-instance tests remain our primary analysis. Within MOMENT models, mean gains are 2.29, 1.43, and 1.76 points for Small, Base, and Large, showing no consistent increase with scale.

For MantisV2 and UniTS, the median differences of 0.80 and 0.40 points are considerably smaller than their mean differences of 2.58 and 2.24 points, respectively. This suggests that their mean improvements are influenced by a subset of datasets with larger positive differences. NuTime shows the largest and most consistent benefit, with a median improvement of 4.20 points and pretraining wins on 21 of 26 datasets. The small standard deviations reported for the random baselines describe variation across encoder initialization seeds after aggregation; they do not represent uncertainty across datasets. The dataset-level results in Appendix~\ref{sec:complete_results} show that pretraining does not improve performance on every dataset. The aggregate differences therefore represent average rather than universal gains, and non-significant Wilcoxon results should not be interpreted as evidence of equivalence.

\subsection{Does the result depend on classifier capacity?}

Table~\ref{tab:classifier_decomposition} decomposes the pretrained--random difference by downstream classifier. Ridge and logistic regression are linear, whereas RBF SVM and random forest allow nonlinear decision boundaries while keeping the encoder frozen.

\begin{table}[htbp]
\centering
\caption{Pretrained minus random-init accuracy ($\Delta$, percentage points) for each downstream classifier. Linear and nonlinear columns report the average difference within each classifier family.}
\label{tab:classifier_decomposition}
\vspace{0.5em}
\begin{tabular}{l c c c c c c}
\toprule
\textbf{Instance} & \textbf{Ridge} & \textbf{LogReg} & \textbf{RBF} & \textbf{RF} & \textbf{Linear} & \textbf{Nonlinear} \\
\midrule
MOMENT-Small & 2.59 & 2.93 & 1.09 & 2.53 & 2.76 & 1.81 \\
MOMENT-Base & 1.69 & 1.49 & 1.47 & 1.09 & 1.59 & 1.28 \\
MOMENT-Large & 2.84 & 2.35 & 1.24 & 0.62 & 2.59 & 0.93 \\
Mantis & 0.38 & 1.92 & 4.71 & 3.23 & 1.15 & 3.97 \\
MantisV2 & 1.33 & 1.55 & 4.26 & 3.19 & 1.44 & 3.73 \\
UniTS & 1.53 & 2.00 & 1.82 & 3.62 & 1.76 & 2.72 \\
NuTime & 4.09 & 5.94 & 4.95 & 5.78 & 5.01 & 5.36 \\
\bottomrule
\end{tabular}
\end{table}

Averaged across model instances, the pretrained advantage is approximately 2.33 percentage points for the linear classifiers and 2.83 points for the nonlinear classifiers. Allowing nonlinear decision boundaries therefore does not show a substantially larger benefit that is consistently hidden from linear classifiers.

The classifier-level results nevertheless show different patterns across model instances. For the MOMENT family, the pretrained advantage is generally smaller under nonlinear classifiers. For MOMENT-Large, for example, the average difference decreases from 2.59 points under the linear classifiers to 0.93 points under the nonlinear classifiers. Mantis and MantisV2 show the opposite pattern, with nonlinear differences of 3.97 and 3.73 points, compared with linear differences of 1.15 and 1.44 points. NuTime shows the most consistent benefit from pretraining, with gains above four percentage points across all four classifiers. Classifier choice changes the gap for individual models, but the average pretrained advantage remains modest under both classifier families.

\subsection{How competitive are time-series-specific methods?}

The architecture-matched comparison measures the contribution of pretrained weights within a fixed foundation-model architecture. A separate question is how the resulting pretrained representations compare with methods developed specifically for time series classification. Table~\ref{tab:common_24} compares the seven pretrained foundation-model instances with ROCKET, Hydra, and Quant on the 24 datasets for which every method has a valid result.

\begin{table}[htbp]
\centering
\caption{Comparison on the common 24 dataset subset. For each foundation model, accuracy is averaged across the four downstream classifiers; ROCKET, Hydra, and Quant are reported using their respective end-to-end classifier outputs. Average rank is computed per dataset across all ten methods, with tied accuracies assigned their average rank. Wins count datasets on which a method attains the highest accuracy; tied methods each receive a win, so the total can exceed 24.}
\label{tab:common_24}
\vspace{0.5em}
\begin{tabular}{l c c c}
\toprule
\textbf{Method} & \textbf{Mean accuracy} & \textbf{Avg. rank} & \textbf{Wins} \\
\midrule
ROCKET       & 0.754 & 3.48 & 8  \\
Hydra        & 0.752 & 3.58 & 10 \\
Quant        & 0.752 & 4.25 & 3  \\
Mantis       & 0.731 & 3.94 & 6  \\
MantisV2     & 0.720 & 4.77 & 4  \\
MOMENT-Small & 0.703 & 5.73 & 3  \\
MOMENT-Large & 0.700 & 6.38 & 0  \\
MOMENT-Base  & 0.698 & 6.67 & 1  \\
NuTime$^\dagger$ & 0.653 & 7.69 & 0  \\
UniTS$^\dagger$  & 0.650 & 8.52 & 1  \\
\bottomrule
\end{tabular}

\vspace{1mm}
\footnotesize
$^\dagger$The model-specific means in Table~\ref{tab:architecture_matched} are 0.626 for NuTime ($n=26$) and 0.651 for UniTS ($n=27$); values here use the common 24 dataset subset.
\end{table}

Under this descriptive averaging protocol, the time-series-specific methods obtain the strongest aggregate results on the common subset. These methods also avoid large-scale pretraining; Appendix~\ref{app:cpu_diagnostic} provides additional efficiency context. ROCKET achieves the highest mean accuracy, at 0.754, and the best average rank, at 3.48. Hydra records the largest number of dataset wins, with 10. Quant obtains a similar mean accuracy of 0.752.

Mantis is the strongest foundation model, with a mean accuracy of 0.731 and an average rank of 3.94. Although its mean accuracy is below those of ROCKET, Hydra, and Quant, Mantis obtains a better average rank than Quant and records six dataset wins. This indicates that the strongest pretrained foundation model remains competitive on a dataset-by-dataset basis, even though the three time-series-specific methods obtain higher mean accuracy. The remaining foundation models rank below all three time-series-specific references on average.

This cross-method comparison is descriptive rather than architecture-matched. In particular, each foundation-model result is averaged across four downstream classifiers, whereas ROCKET, Hydra, and Quant are represented by their end-to-end classifier outputs. Nevertheless, the comparison provides an external reference for the level of performance achieved by the frozen foundation-model representations.

These results should not be interpreted as showing that foundation models are universally inferior to time-series-specific methods. The evaluation is restricted to frozen MTSC representations and does not assess full fine-tuning, forecasting, anomaly detection, or other downstream tasks. Under the evaluated setting, however, the improvements provided by pretraining do not result in higher average accuracy than the strongest time-series-specific methods. Appendix~\ref{app:cpu_diagnostic} shows that within MOMENT, latency rises from 82.48\,ms to 1354.18\,ms from Small to Large, while the pretrained-over-random gain does not increase (2.29 vs.\ 1.76 points); this synthetic diagnostic is not a full cost analysis.

\subsection{Discussion and limitations}

The results show that randomly initialized foundation-model architectures can produce useful frozen representations even without pretraining. One possible explanation is that high-dimensional randomly initialized networks can act as useful feature mappings~\citep{rahimi2007,jacot2018}. ROCKET~\citep{rocket} provides a related example in time series classification, where randomly parameterized convolutional kernels produce effective features. Our results are therefore consistent with the possibility that part of the downstream performance of foundation models comes from architectural properties that are already present before pretraining.

A second possible contributor is the multivariate inference pipeline. The evaluated foundation models largely process channels independently and then combine the resulting representations through model-specific aggregation operations~\citep{pinar2025fm4mtsc}. Since the same architecture, preprocessing, and aggregation procedure are used under pretrained and random initialization, these components can contribute to classification performance independently of the learned weights. The classifier-specific results also show that the relatively small average difference cannot be explained only by limited linear accessibility: nonlinear classifiers change the magnitude of the difference for individual models but do not show a consistently large advantage for pretraining. These observations do not imply that pretrained representations are equivalent to randomly initialized representations. Pretraining achieves higher mean accuracy for every model instance, and NuTime retains a statistically significant advantage after multiple-comparison correction. Rather, the results show that the contribution of pretrained weights should be measured relative to an architecture-matched random baseline instead of being inferred from the absolute performance of a pretrained model alone.

Our study has several limitations. First, the conclusions apply only to frozen representation evaluation for MTSC and should not be extended to full fine-tuning or other downstream tasks. Fine-tuning may make use of pretrained information that is not directly accessible through frozen representations. Second, dataset coverage differs across model instances because some model-dataset combinations could not be completed due to computational or pipeline constraints. We address this by restricting each pretrained and random comparison to paired datasets and by using a common 24 dataset subset for the comparison with the time-series-specific methods. Third, the comparison with ROCKET, Hydra, and Quant is not a fully matched comparison. The foundation models are summarized across four downstream classifiers, whereas the time-series-specific methods produce end-to-end predictions using their own classification pipelines. The comparison should therefore be interpreted as an external performance reference rather than a controlled test of the foundation-model architecture. Finally, although large-scale pretraining motivates the resource question in this work, we do not measure pretraining time, energy consumption, or carbon emissions. The controlled CPU diagnostic in Appendix~\ref{app:cpu_diagnostic} provides limited inference-side context, including a substantial increase in latency and registered operations with MOMENT scale, but it does not provide dataset-native end-to-end latency or a complete cost-benefit analysis. Our primary results therefore quantify the downstream accuracy associated with pretrained weights rather than the full computational cost of obtaining and deploying them.
\section{Conclusion}
\label{sec:conclusion}

We systematically evaluated seven instances from four time series foundation model families by comparing their released pretrained weights against architecture-matched random initialization on the UEA MTSC archive. Across four downstream classifiers spanning linear and nonlinear decision functions and five random initialization seeds, pretraining improves mean accuracy for all seven model instances, but by a modest 1.43-5.19 percentage points per instance and 2.58 points overall across 183 paired dataset-level comparisons. After Holm correction, only NuTime retains a statistically significant per-instance advantage. The pooled comparison is also significant ($p<0.001$), although the Holm-corrected per-instance tests remain our primary evidence and the average pretrained--random gap remains modest across both classifier families.

Our comparison with time-series-specific methods provides a complementary perspective: ROCKET, Hydra, and Quant remain competitive with the evaluated pretrained foundation models without requiring large-scale pretraining. Together, these results suggest that MTSC evaluations should distinguish the contribution of pretrained weights from that of the underlying architecture and representation pipeline, and should assess this contribution relative to strong time-series-specific alternatives. While full fine-tuning may show benefits not captured by frozen representations, our findings support the inclusion of architecture-matched random initialization and non-pretrained time-series methods as standard controls when evaluating the downstream value of large-scale time series pretraining.
\begin{ack}
This publication was conducted with the financial support of Taighde Éireann – Research Ireland under Grant [ML-Labs 18/CRT/6183]. For the purpose of Open Access, the author has applied a CC BY public copyright license to any Author Accepted Manuscript version arising from this submission.
\end{ack}
\bibliographystyle{plainnat}
\bibliography{references}

\appendix

\appendix

\section{Dataset coverage}
\label{app:dataset_coverage}

A model-dataset result is included in the aggregate foundation model analysis
only when ridge classification, logistic regression, RBF SVM, and random
forest produce valid results under both the pretrained and randomly initialized
conditions. Table~\ref{tab:dataset_coverage} summarizes the datasets for which
at least one foundation-model instance does not have a complete four classifier
result.

\begin{table*}[h]
\centering
\caption{Foundation model dataset coverage. ``Missing'' indicates that a
complete four classifier result is unavailable under at least one of the
pretrained or randomly initialized conditions.}
\label{tab:dataset_coverage}
\small
\begin{tabular}{l p{0.72\textwidth}}
\toprule
\textbf{Dataset} & \textbf{Coverage in the four classifier aggregate} \\
\midrule
DuckDuckGeese
& Missing for UniTS; complete for the other six model instances. \\

FaceDetection
& Complete only for UniTS; missing for all MOMENT instances, Mantis,
MantisV2, and NuTime. \\

InsectWingbeat
& Not evaluated for any of the seven foundation model instances. \\

PenDigits
& Missing for NuTime; complete for the other six model instances. \\

PEMS-SF
& Missing for all seven foundation model instances. \\

StandWalkJump
& Complete for UniTS and NuTime. For the three MOMENT instances, Mantis,
and MantisV2, the logistic-regression result is unavailable, so no
four-classifier aggregate is reported. \\
\bottomrule
\end{tabular}
\end{table*}

The common 24 dataset subset used for the cross-method comparison therefore
consists of the complete UEA archive excluding DuckDuckGeese, FaceDetection,
InsectWingbeat, PenDigits, PEMS-SF, and StandWalkJump.

\section{Implementation details}
\label{sec:implementation}

For each foundation model instance, we use the released pretrained checkpoint
and the corresponding official implementation. The downstream
classifier selection protocol used throughout the experiments is detailed
below.

All downstream hyperparameters are selected using three-fold stratified
cross-validation on the official UEA training split. The folds are shuffled
with \texttt{random\_state=42}, classification accuracy is used for selection,
and the selected pipeline is refitted on the complete training split before a
single evaluation on the official test split. Standardization is included
inside the cross-validation pipeline for ridge classification, logistic
regression, and the RBF SVM.

\begin{table*}[h]
\centering
\caption{Downstream-classifier configurations. The same grids and
cross-validation procedure are used for every dataset, model instance,
pretraining condition, and random encoder seed.}
\label{tab:classifier_grids}
\small
\begin{tabular}{l p{0.35\textwidth} p{0.43\textwidth}}
\toprule
\textbf{Classifier} & \textbf{Fixed configuration} &
\textbf{Selected hyperparameters} \\
\midrule
Ridge & \texttt{StandardScaler}; \texttt{RidgeClassifier} &
$\alpha \in \{10^{-3},10^{-2},10^{-1},1,10,10^{2},10^{3}\}$ \\
Logistic regression & \texttt{StandardScaler}; \texttt{max\_iter=5000};
\texttt{solver=lbfgs} &
$C \in \{10^{-3},10^{-2},10^{-1},1,10,10^{2},10^{3}\}$ \\
RBF SVM & \texttt{StandardScaler}; \texttt{kernel=rbf} &
$C \in \{10^{-2},10^{-1},1,10,10^{2}\}$;
\texttt{gamma} $\in \{\texttt{scale},\texttt{auto}\}$ \\
Random forest & 200 trees; \texttt{max\_features=sqrt};
\texttt{random\_state=42} &
\texttt{max\_depth} $\in \{\texttt{None},10\}$;
\texttt{min\_samples\_leaf} $\in \{1,2\}$ \\
\bottomrule
\end{tabular}
\end{table*}

For each randomly initialized encoder seed, downstream model selection is
performed independently using this same training-only procedure. No classifier
or hyperparameter is selected using the UEA test split.

\section{Per-dataset foundation-model results}
\label{sec:complete_results}

This appendix reports the per-dataset results underlying the aggregate foundation-model comparisons in the main paper. All 30 UEA datasets are listed; unavailable model-dataset results are shown as dashes. Dataset
coverage and the construction of the common 24 dataset subset are described in Appendix~\ref{app:dataset_coverage}.

For random initialization, classifier results are additionally averaged across the five initialization seeds $\{0, 7, 42, 123, 2026\}$. Tables~\ref{tab:random_ridge}-\ref{tab:random_rf} and Tables~\ref{tab:pretrained_ridge}-\ref{tab:pretrained_rf} provide the classifier specific results.

Dataset coverage is not uniform across foundation model instances or conditions. The mean at the bottom of each aggregate table is computed over entries with complete results for all four downstream classifiers. All pretrained-random comparisons in Table~\ref{tab:architecture_matched} are restricted to datasets available under both conditions for the corresponding model instance. These paired dataset-level values are also used to compute the mean and median differences and the win/tie/loss counts reported in Table~\ref{tab:architecture_matched}.

\begin{table}[htbp]
\centering
\caption{Per-dataset classification accuracy of randomly initialized foundation models, averaged across all four downstream classifiers and five initialization seeds. Entries with any missing classifier result are shown as a dash.}
\label{tab:random_aggregate}
\vspace{0.5em}
\scriptsize
\begin{tabular}{l c c c c c c c}
\toprule
\textbf{Dataset} & \textbf{MOMENT-S} & \textbf{MOMENT-B} & \textbf{MOMENT-L} & \textbf{Mantis} & \textbf{MantisV2} & \textbf{UniTS} & \textbf{NuTime} \\
\midrule
Articulary WordRecognition & 0.983 & 0.983 & 0.983 & 0.988 & 0.984 & 0.899 & 0.903 \\
Atrial Fibrillation       & 0.310 & 0.330 & 0.387 & 0.290 & 0.227 & 0.320 & 0.240 \\
BasicMotions              & 0.982 & 0.986 & 0.981 & 0.993 & 0.994 & 0.781 & 0.482 \\
Character Trajectories    & 0.977 & 0.980 & 0.979 & 0.944 & 0.938 & 0.950 & 0.828 \\
Cricket                   & 0.938 & 0.928 & 0.940 & 0.974 & 0.979 & 0.848 & 0.923 \\
DuckDuckGeese             & 0.266 & 0.302 & 0.312 & 0.550 & 0.524 & --    & 0.249 \\
Eigen Worms               & 0.525 & 0.522 & 0.514 & 0.652 & 0.644 & 0.538 & 0.704 \\
Epilepsy                  & 0.934 & 0.933 & 0.928 & 0.974 & 0.985 & 0.853 & 0.845 \\
ERing                     & 0.909 & 0.928 & 0.922 & 0.905 & 0.906 & 0.855 & 0.708 \\
EthanolConcentration      & 0.296 & 0.299 & 0.293 & 0.404 & 0.410 & 0.251 & 0.317 \\
Face Detection            & --    & --    & --    & --    & --    & 0.567 & --    \\
FingerMovements           & 0.572 & 0.558 & 0.567 & 0.531 & 0.514 & 0.506 & 0.513 \\
HandMovementDirection     & 0.488 & 0.512 & 0.522 & 0.362 & 0.320 & 0.392 & 0.251 \\
Handwriting               & 0.158 & 0.172 & 0.173 & 0.229 & 0.243 & 0.179 & 0.115 \\
Heartbeat                 & 0.681 & 0.677 & 0.672 & 0.740 & 0.724 & 0.690 & 0.682 \\
Japanese Vowels           & 0.913 & 0.913 & 0.889 & 0.794 & 0.780 & 0.823 & 0.766 \\
Libras                    & 0.814 & 0.796 & 0.788 & 0.821 & 0.818 & 0.721 & 0.696 \\
LSST                      & 0.558 & 0.565 & 0.569 & 0.661 & 0.665 & 0.462 & 0.453 \\
MotorImagery              & 0.504 & 0.504 & 0.510 & 0.515 & 0.526 & 0.495 & 0.552 \\
NATOPS                    & 0.879 & 0.884 & 0.884 & 0.811 & 0.807 & 0.607 & 0.594 \\
PenDigits                 & 0.972 & 0.972 & 0.972 & 0.959 & 0.960 & 0.972 & --    \\
PEMS-SF                   & --    & --    & --    & --    & --    & --    & --    \\
PhonemeSpectra            & 0.124 & 0.130 & 0.128 & 0.219 & 0.218 & 0.103 & 0.102 \\
RacketSports              & 0.705 & 0.701 & 0.688 & 0.819 & 0.800 & 0.629 & 0.725 \\
SelfRegulationSCP1        & 0.830 & 0.840 & 0.830 & 0.842 & 0.842 & 0.831 & 0.777 \\
SelfRegulationSCP2        & 0.549 & 0.542 & 0.539 & 0.528 & 0.528 & 0.499 & 0.523 \\
SpokenArabicDigits        & 0.950 & 0.957 & 0.961 & 0.930 & 0.928 & 0.916 & 0.842 \\
StandWalkJump             & --    & --    & --    & --    & --    & 0.460 & 0.373 \\
UWaveGestureLibrary       & 0.868 & 0.858 & 0.861 & 0.826 & 0.800 & 0.822 & 0.753 \\
\midrule
Mean                      & 0.680 & 0.684 & 0.684 & 0.702 & 0.695 & 0.628 & 0.574 \\
\bottomrule
\end{tabular}
\end{table}

\begin{table}[htbp]
\centering
\caption{Per-dataset classification accuracy of pretrained foundation models, averaged across all four downstream classifiers. Entries with any missing classifier result are shown as a dash.}
\label{tab:pretrained_aggregate}
\vspace{0.5em}
\scriptsize
\begin{tabular}{l c c c c c c c}
\toprule
\textbf{Dataset} & \textbf{MOMENT-S} & \textbf{MOMENT-B} & \textbf{MOMENT-L} & \textbf{Mantis} & \textbf{MantisV2} & \textbf{UniTS} & \textbf{NuTime} \\
\midrule
Articulary WordRecognition & 0.992 & 0.989 & 0.990 & 0.993 & 0.993 & 0.903 & 0.983 \\
Atrial Fibrillation       & 0.117 & 0.150 & 0.183 & 0.200 & 0.133 & 0.383 & 0.317 \\
BasicMotions              & 1.000 & 1.000 & 0.988 & 1.000 & 1.000 & 0.888 & 0.731 \\
Character Trajectories    & 0.977 & 0.975 & 0.975 & 0.971 & 0.962 & 0.940 & 0.934 \\
Cricket                   & 0.983 & 0.986 & 0.983 & 0.993 & 0.986 & 0.830 & 0.972 \\
DuckDuckGeese             & 0.455 & 0.440 & 0.475 & 0.430 & 0.495 & --    & 0.195 \\
Eigen Worms               & 0.740 & 0.704 & 0.687 & 0.761 & 0.807 & 0.603 & 0.691 \\
Epilepsy                  & 1.000 & 0.996 & 0.993 & 0.996 & 1.000 & 0.944 & 0.884 \\
ERing                     & 0.937 & 0.903 & 0.874 & 0.955 & 0.968 & 0.864 & 0.871 \\
EthanolConcentration      & 0.318 & 0.298 & 0.318 & 0.293 & 0.391 & 0.265 & 0.438 \\
Face Detection            & --    & --    & --    & --    & --    & 0.566 & --    \\
FingerMovements           & 0.603 & 0.568 & 0.547 & 0.583 & 0.570 & 0.508 & 0.522 \\
HandMovementDirection     & 0.270 & 0.260 & 0.304 & 0.253 & 0.247 & 0.412 & 0.321 \\
Handwriting               & 0.238 & 0.269 & 0.231 & 0.369 & 0.301 & 0.200 & 0.134 \\
Heartbeat                 & 0.732 & 0.721 & 0.722 & 0.802 & 0.801 & 0.674 & 0.707 \\
Japanese Vowels           & 0.947 & 0.943 & 0.936 & 0.973 & 0.963 & 0.868 & 0.786 \\
Libras                    & 0.864 & 0.864 & 0.861 & 0.875 & 0.899 & 0.717 & 0.774 \\
LSST                      & 0.584 & 0.588 & 0.603 & 0.648 & 0.658 & 0.462 & 0.413 \\
MotorImagery              & 0.505 & 0.460 & 0.487 & 0.540 & 0.512 & 0.505 & 0.535 \\
NATOPS                    & 0.899 & 0.888 & 0.906 & 0.932 & 0.883 & 0.704 & 0.683 \\
PenDigits                 & 0.953 & 0.960 & 0.967 & 0.959 & 0.961 & 0.965 & --    \\
PEMS-SF                   & --    & --    & --    & --    & --    & --    & --    \\
PhonemeSpectra            & 0.212 & 0.208 & 0.214 & 0.296 & 0.302 & 0.105 & 0.109 \\
RacketSports              & 0.794 & 0.840 & 0.808 & 0.931 & 0.908 & 0.745 & 0.724 \\
SelfRegulationSCP1        & 0.834 & 0.810 & 0.817 & 0.824 & 0.797 & 0.802 & 0.822 \\
SelfRegulationSCP2        & 0.489 & 0.487 & 0.537 & 0.504 & 0.521 & 0.501 & 0.533 \\
SpokenArabicDigits        & 0.955 & 0.961 & 0.936 & 0.964 & 0.898 & 0.943 & 0.894 \\
StandWalkJump             & --    & --    & --    & --    & --    & 0.450 & 0.400 \\
UWaveGestureLibrary       & 0.881 & 0.874 & 0.905 & 0.879 & 0.780 & 0.824 & 0.891 \\
\midrule
Mean                      & 0.703 & 0.698 & 0.702 & 0.728 & 0.721 & 0.651 & 0.626 \\
\bottomrule
\end{tabular}
\end{table}


\begin{table}[htbp]
\centering
\caption{Per-dataset top-one accuracy of randomly initialized foundation models using ridge classification. Values are averaged across five initialization seeds.}
\label{tab:random_ridge}
\vspace{0.5em}
\scriptsize
\begin{tabular}{l c c c c c c c}
\toprule
\textbf{Dataset} & \textbf{MOMENT-S} & \textbf{MOMENT-B} & \textbf{MOMENT-L} & \textbf{Mantis} & \textbf{MantisV2} & \textbf{UniTS} & \textbf{NuTime} \\
\midrule
Articulary WordRecognition & 0.983 & 0.983 & 0.982 & 0.997 & 0.995 & 0.954 & 0.911 \\
Atrial Fibrillation       & 0.293 & 0.360 & 0.413 & 0.267 & 0.240 & 0.307 & 0.267 \\
BasicMotions              & 0.995 & 0.995 & 0.985 & 1.000 & 1.000 & 0.820 & 0.455 \\
Character Trajectories    & 0.981 & 0.983 & 0.981 & 0.982 & 0.978 & 0.967 & 0.865 \\
Cricket                   & 0.942 & 0.928 & 0.939 & 0.992 & 0.997 & 0.822 & 0.975 \\
DuckDuckGeese             & 0.304 & 0.348 & 0.340 & 0.616 & 0.580 & --    & 0.264 \\
Eigen Worms               & 0.531 & 0.510 & 0.490 & 0.638 & 0.641 & 0.527 & 0.756 \\
Epilepsy                  & 0.926 & 0.928 & 0.929 & 0.983 & 0.993 & 0.841 & 0.887 \\
ERing                     & 0.901 & 0.919 & 0.912 & 0.930 & 0.928 & 0.860 & 0.713 \\
EthanolConcentration      & 0.319 & 0.326 & 0.321 & 0.453 & 0.463 & 0.246 & 0.315 \\
Face Detection            & --    & --    & --    & --    & --    & 0.578 & --    \\
FingerMovements           & 0.586 & 0.552 & 0.588 & 0.568 & 0.560 & 0.532 & 0.496 \\
HandMovementDirection     & 0.527 & 0.587 & 0.600 & 0.389 & 0.346 & 0.395 & 0.262 \\
Handwriting               & 0.197 & 0.221 & 0.220 & 0.235 & 0.246 & 0.179 & 0.108 \\
Heartbeat                 & 0.639 & 0.634 & 0.624 & 0.715 & 0.696 & 0.691 & 0.642 \\
Japanese Vowels           & 0.922 & 0.906 & 0.886 & 0.885 & 0.856 & 0.891 & 0.772 \\
Libras                    & 0.796 & 0.767 & 0.752 & 0.871 & 0.879 & 0.676 & 0.652 \\
LSST                      & 0.555 & 0.556 & 0.558 & 0.652 & 0.658 & 0.463 & 0.443 \\
MotorImagery              & 0.516 & 0.506 & 0.518 & 0.590 & 0.592 & 0.466 & 0.552 \\
NATOPS                    & 0.908 & 0.903 & 0.901 & 0.817 & 0.828 & 0.666 & 0.589 \\
PenDigits                 & 0.969 & 0.970 & 0.970 & 0.962 & 0.963 & 0.980 & --    \\
PEMS-SF                   & --    & --    & --    & --    & --    & --    & --    \\
PhonemeSpectra            & 0.120 & 0.129 & 0.125 & 0.228 & 0.229 & 0.102 & 0.107 \\
RacketSports              & 0.726 & 0.717 & 0.707 & 0.841 & 0.838 & 0.667 & 0.717 \\
SelfRegulationSCP1        & 0.842 & 0.853 & 0.842 & 0.853 & 0.868 & 0.847 & 0.808 \\
SelfRegulationSCP2        & 0.553 & 0.538 & 0.539 & 0.520 & 0.516 & 0.498 & 0.539 \\
SpokenArabicDigits        & 0.965 & 0.971 & 0.975 & 0.987 & 0.987 & 0.960 & 0.894 \\
StandWalkJump             & 0.387 & 0.320 & 0.360 & 0.400 & 0.347 & 0.547 & 0.400 \\
UWaveGestureLibrary       & 0.866 & 0.857 & 0.863 & 0.919 & 0.893 & 0.861 & 0.836 \\
\midrule
Mean                      & 0.676 & 0.676 & 0.678 & 0.714 & 0.708 & 0.642 & 0.586 \\
\bottomrule
\end{tabular}
\end{table}

\begin{table}[htbp]
\centering
\caption{Per-dataset top-one accuracy of randomly initialized foundation models using logistic regression. Values are averaged across five initialization seeds.}
\label{tab:random_logreg}
\vspace{0.5em}
\scriptsize
\begin{tabular}{l c c c c c c c}
\toprule
\textbf{Dataset} & \textbf{MOMENT-S} & \textbf{MOMENT-B} & \textbf{MOMENT-L} & \textbf{Mantis} & \textbf{MantisV2} & \textbf{UniTS} & \textbf{NuTime} \\
\midrule
Articulary WordRecognition & 0.987 & 0.985 & 0.985 & 0.988 & 0.987 & 0.916 & 0.897 \\
Atrial Fibrillation       & 0.293 & 0.373 & 0.400 & 0.293 & 0.227 & 0.333 & 0.200 \\
BasicMotions              & 0.990 & 0.995 & 0.970 & 1.000 & 1.000 & 0.790 & 0.495 \\
Character Trajectories    & 0.980 & 0.983 & 0.982 & 0.973 & 0.968 & 0.952 & 0.848 \\
Cricket                   & 0.939 & 0.936 & 0.944 & 0.972 & 0.975 & 0.856 & 0.936 \\
DuckDuckGeese             & 0.260 & 0.272 & 0.276 & 0.596 & 0.548 & --    & 0.272 \\
Eigen Worms               & 0.530 & 0.524 & 0.495 & 0.710 & 0.690 & 0.560 & 0.705 \\
Epilepsy                  & 0.938 & 0.933 & 0.932 & 0.990 & 0.999 & 0.842 & 0.854 \\
ERing                     & 0.921 & 0.935 & 0.933 & 0.887 & 0.896 & 0.856 & 0.695 \\
EthanolConcentration      & 0.339 & 0.352 & 0.344 & 0.438 & 0.477 & 0.257 & 0.376 \\
Face Detection            & --    & --    & --    & --    & --    & 0.576 & --    \\
FingerMovements           & 0.588 & 0.580 & 0.592 & 0.570 & 0.550 & 0.478 & 0.506 \\
HandMovementDirection     & 0.516 & 0.600 & 0.592 & 0.411 & 0.368 & 0.403 & 0.249 \\
Handwriting               & 0.189 & 0.200 & 0.191 & 0.274 & 0.297 & 0.185 & 0.110 \\
Heartbeat                 & 0.663 & 0.650 & 0.642 & 0.725 & 0.713 & 0.651 & 0.645 \\
Japanese Vowels           & 0.926 & 0.935 & 0.907 & 0.783 & 0.774 & 0.858 & 0.767 \\
Libras                    & 0.840 & 0.836 & 0.833 & 0.841 & 0.850 & 0.722 & 0.691 \\
LSST                      & 0.554 & 0.550 & 0.556 & 0.667 & 0.670 & 0.494 & 0.456 \\
MotorImagery              & 0.514 & 0.508 & 0.516 & 0.580 & 0.564 & 0.506 & 0.548 \\
NATOPS                    & 0.911 & 0.918 & 0.916 & 0.809 & 0.808 & 0.641 & 0.574 \\
PenDigits                 & 0.974 & 0.971 & 0.971 & 0.966 & 0.968 & 0.968 & --    \\
PEMS-SF                   & --    & --    & --    & --    & --    & --    & --    \\
PhonemeSpectra            & 0.121 & 0.126 & 0.122 & 0.227 & 0.232 & 0.103 & 0.111 \\
RacketSports              & 0.726 & 0.714 & 0.708 & 0.809 & 0.762 & 0.641 & 0.712 \\
SelfRegulationSCP1        & 0.822 & 0.841 & 0.820 & 0.852 & 0.876 & 0.817 & 0.791 \\
SelfRegulationSCP2        & 0.549 & 0.533 & 0.527 & 0.477 & 0.519 & 0.493 & 0.487 \\
SpokenArabicDigits        & 0.959 & 0.966 & 0.965 & 0.966 & 0.968 & 0.935 & 0.828 \\
StandWalkJump             & --    & --    & --    & --    & --    & 0.493 & 0.360 \\
UWaveGestureLibrary       & 0.852 & 0.837 & 0.842 & 0.824 & 0.790 & 0.821 & 0.770 \\
\midrule
Mean                      & 0.688 & 0.694 & 0.691 & 0.716 & 0.710 & 0.635 & 0.572 \\
\bottomrule
\end{tabular}
\end{table}

\begin{table}[htbp]
\centering
\caption{Per-dataset top-one accuracy of randomly initialized foundation models using RBF SVM. Values are averaged across five initialization seeds.}
\label{tab:random_svm}
\vspace{0.5em}
\scriptsize
\begin{tabular}{l c c c c c c c}
\toprule
\textbf{Dataset} & \textbf{MOMENT-S} & \textbf{MOMENT-B} & \textbf{MOMENT-L} & \textbf{Mantis} & \textbf{MantisV2} & \textbf{UniTS} & \textbf{NuTime} \\
\midrule
Articulary WordRecognition & 0.985 & 0.983 & 0.984 & 0.979 & 0.976 & 0.894 & 0.908 \\
Atrial Fibrillation       & 0.307 & 0.293 & 0.307 & 0.373 & 0.240 & 0.267 & 0.253 \\
BasicMotions              & 0.980 & 0.995 & 0.995 & 0.990 & 0.975 & 0.780 & 0.510 \\
Character Trajectories    & 0.974 & 0.979 & 0.977 & 0.911 & 0.907 & 0.956 & 0.832 \\
Cricket                   & 0.939 & 0.933 & 0.950 & 0.972 & 0.978 & 0.856 & 0.942 \\
DuckDuckGeese             & 0.272 & 0.276 & 0.304 & 0.412 & 0.428 & --    & 0.256 \\
Eigen Worms               & 0.502 & 0.516 & 0.534 & 0.608 & 0.597 & 0.533 & 0.684 \\
Epilepsy                  & 0.955 & 0.943 & 0.945 & 0.962 & 0.975 & 0.871 & 0.820 \\
ERing                     & 0.913 & 0.922 & 0.914 & 0.897 & 0.889 & 0.848 & 0.714 \\
EthanolConcentration      & 0.269 & 0.261 & 0.259 & 0.369 & 0.354 & 0.256 & 0.287 \\
Face Detection            & --    & --    & --    & --    & --    & 0.566 & --    \\
FingerMovements           & 0.538 & 0.548 & 0.536 & 0.470 & 0.458 & 0.504 & 0.532 \\
HandMovementDirection     & 0.451 & 0.408 & 0.443 & 0.300 & 0.295 & 0.384 & 0.241 \\
Handwriting               & 0.102 & 0.110 & 0.113 & 0.184 & 0.184 & 0.185 & 0.127 \\
Heartbeat                 & 0.721 & 0.721 & 0.722 & 0.750 & 0.745 & 0.707 & 0.722 \\
Japanese Vowels           & 0.906 & 0.912 & 0.890 & 0.734 & 0.724 & 0.806 & 0.768 \\
Libras                    & 0.776 & 0.748 & 0.739 & 0.786 & 0.764 & 0.779 & 0.752 \\
LSST                      & 0.585 & 0.596 & 0.600 & 0.666 & 0.672 & 0.454 & 0.468 \\
MotorImagery              & 0.492 & 0.502 & 0.498 & 0.442 & 0.470 & 0.518 & 0.556 \\
NATOPS                    & 0.836 & 0.832 & 0.836 & 0.830 & 0.824 & 0.572 & 0.611 \\
PenDigits                 & 0.979 & 0.978 & 0.978 & 0.963 & 0.964 & 0.976 & --    \\
PEMS-SF                   & --    & --    & --    & --    & --    & --    & --    \\
PhonemeSpectra            & 0.134 & 0.138 & 0.138 & 0.211 & 0.208 & 0.110 & 0.101 \\
RacketSports              & 0.689 & 0.689 & 0.682 & 0.818 & 0.832 & 0.653 & 0.766 \\
SelfRegulationSCP1        & 0.838 & 0.840 & 0.838 & 0.837 & 0.828 & 0.823 & 0.762 \\
SelfRegulationSCP2        & 0.543 & 0.551 & 0.548 & 0.578 & 0.538 & 0.503 & 0.560 \\
SpokenArabicDigits        & 0.961 & 0.965 & 0.966 & 0.908 & 0.895 & 0.926 & 0.848 \\
StandWalkJump             & 0.373 & 0.347 & 0.373 & 0.400 & 0.373 & 0.413 & 0.440 \\
UWaveGestureLibrary       & 0.894 & 0.885 & 0.894 & 0.800 & 0.782 & 0.829 & 0.706 \\
\midrule
Mean                      & 0.663 & 0.662 & 0.665 & 0.672 & 0.662 & 0.628 & 0.583 \\
\bottomrule
\end{tabular}
\end{table}

\begin{table}[htbp]
\centering
\caption{Per-dataset top-one accuracy of randomly initialized foundation models using random forest. Values are averaged across five initialization seeds.}
\label{tab:random_rf}
\vspace{0.5em}
\scriptsize
\begin{tabular}{l c c c c c c c}
\toprule
\textbf{Dataset} & \textbf{MOMENT-S} & \textbf{MOMENT-B} & \textbf{MOMENT-L} & \textbf{Mantis} & \textbf{MantisV2} & \textbf{UniTS} & \textbf{NuTime} \\
\midrule
Articulary WordRecognition & 0.977 & 0.982 & 0.980 & 0.989 & 0.976 & 0.831 & 0.895 \\
Atrial Fibrillation       & 0.347 & 0.293 & 0.427 & 0.227 & 0.200 & 0.373 & 0.240 \\
BasicMotions              & 0.965 & 0.960 & 0.975 & 0.980 & 1.000 & 0.735 & 0.470 \\
Character Trajectories    & 0.971 & 0.975 & 0.976 & 0.910 & 0.898 & 0.923 & 0.765 \\
Cricket                   & 0.933 & 0.914 & 0.928 & 0.961 & 0.967 & 0.858 & 0.839 \\
DuckDuckGeese             & 0.228 & 0.312 & 0.328 & 0.576 & 0.540 & --    & 0.204 \\
Eigen Worms               & 0.536 & 0.537 & 0.537 & 0.652 & 0.647 & 0.533 & 0.670 \\
Epilepsy                  & 0.916 & 0.928 & 0.906 & 0.959 & 0.974 & 0.857 & 0.820 \\
ERing                     & 0.902 & 0.936 & 0.930 & 0.906 & 0.910 & 0.856 & 0.709 \\
EthanolConcentration      & 0.259 & 0.256 & 0.249 & 0.357 & 0.346 & 0.243 & 0.288 \\
Face Detection            & --    & --    & --    & --    & --    & 0.549 & --    \\
FingerMovements           & 0.574 & 0.550 & 0.550 & 0.516 & 0.488 & 0.508 & 0.518 \\
HandMovementDirection     & 0.457 & 0.454 & 0.454 & 0.349 & 0.273 & 0.386 & 0.254 \\
Handwriting               & 0.143 & 0.157 & 0.167 & 0.223 & 0.245 & 0.166 & 0.115 \\
Heartbeat                 & 0.702 & 0.704 & 0.701 & 0.769 & 0.742 & 0.713 & 0.720 \\
Japanese Vowels           & 0.897 & 0.899 & 0.872 & 0.774 & 0.765 & 0.736 & 0.758 \\
Libras                    & 0.846 & 0.836 & 0.826 & 0.786 & 0.779 & 0.709 & 0.689 \\
LSST                      & 0.537 & 0.557 & 0.562 & 0.656 & 0.661 & 0.439 & 0.443 \\
MotorImagery              & 0.496 & 0.500 & 0.506 & 0.450 & 0.478 & 0.490 & 0.552 \\
NATOPS                    & 0.863 & 0.881 & 0.886 & 0.788 & 0.769 & 0.551 & 0.600 \\
PenDigits                 & 0.968 & 0.969 & 0.970 & 0.946 & 0.945 & 0.966 & --    \\
PEMS-SF                   & --    & --    & --    & --    & --    & --    & --    \\
PhonemeSpectra            & 0.121 & 0.126 & 0.128 & 0.210 & 0.205 & 0.096 & 0.088 \\
RacketSports              & 0.676 & 0.683 & 0.655 & 0.808 & 0.768 & 0.557 & 0.705 \\
SelfRegulationSCP1        & 0.819 & 0.825 & 0.818 & 0.826 & 0.797 & 0.837 & 0.745 \\
SelfRegulationSCP2        & 0.550 & 0.544 & 0.541 & 0.538 & 0.539 & 0.500 & 0.507 \\
SpokenArabicDigits        & 0.916 & 0.928 & 0.937 & 0.857 & 0.865 & 0.842 & 0.799 \\
StandWalkJump             & 0.440 & 0.427 & 0.413 & 0.440 & 0.360 & 0.387 & 0.293 \\
UWaveGestureLibrary       & 0.860 & 0.852 & 0.848 & 0.761 & 0.736 & 0.775 & 0.698 \\
\midrule
Mean                      & 0.663 & 0.666 & 0.669 & 0.675 & 0.662 & 0.608 & 0.553 \\
\bottomrule
\end{tabular}
\end{table}


\begin{table}[htbp]
\centering
\caption{Per-dataset top-one accuracy of pretrained foundation models using ridge classification.}
\label{tab:pretrained_ridge}
\vspace{0.5em}
\scriptsize
\begin{tabular}{l c c c c c c c}
\toprule
\textbf{Dataset} & \textbf{MOMENT-S} & \textbf{MOMENT-B} & \textbf{MOMENT-L} & \textbf{Mantis} & \textbf{MantisV2} & \textbf{UniTS} & \textbf{NuTime} \\
\midrule
Articulary WordRecognition & 0.997 & 0.990 & 0.997 & 0.993 & 0.993 & 0.933 & 0.990 \\
Atrial Fibrillation       & 0.133 & 0.133 & 0.200 & 0.133 & 0.200 & 0.400 & 0.333 \\
BasicMotions              & 1.000 & 1.000 & 1.000 & 1.000 & 1.000 & 0.900 & 0.750 \\
Character Trajectories    & 0.981 & 0.983 & 0.986 & 0.988 & 0.975 & 0.948 & 0.934 \\
Cricket                   & 1.000 & 0.986 & 1.000 & 0.986 & 0.986 & 0.833 & 0.986 \\
DuckDuckGeese             & 0.480 & 0.520 & 0.580 & 0.440 & 0.560 & --    & 0.180 \\
Eigen Worms               & 0.733 & 0.695 & 0.710 & 0.786 & 0.840 & 0.641 & 0.672 \\
Epilepsy                  & 1.000 & 1.000 & 1.000 & 1.000 & 1.000 & 0.928 & 0.899 \\
ERing                     & 0.937 & 0.918 & 0.885 & 0.952 & 0.970 & 0.848 & 0.889 \\
EthanolConcentration      & 0.304 & 0.308 & 0.323 & 0.297 & 0.407 & 0.243 & 0.483 \\
Face Detection            & --    & --    & --    & --    & --    & 0.585 & --    \\
FingerMovements           & 0.610 & 0.570 & 0.540 & 0.600 & 0.580 & 0.500 & 0.440 \\
HandMovementDirection     & 0.297 & 0.311 & 0.351 & 0.243 & 0.270 & 0.419 & 0.351 \\
Handwriting               & 0.282 & 0.325 & 0.294 & 0.395 & 0.326 & 0.212 & 0.122 \\
Heartbeat                 & 0.737 & 0.707 & 0.707 & 0.810 & 0.800 & 0.649 & 0.688 \\
Japanese Vowels           & 0.951 & 0.932 & 0.941 & 0.981 & 0.965 & 0.900 & 0.792 \\
Libras                    & 0.867 & 0.867 & 0.867 & 0.861 & 0.917 & 0.728 & 0.744 \\
LSST                      & 0.578 & 0.577 & 0.595 & 0.663 & 0.659 & 0.460 & 0.406 \\
MotorImagery              & 0.500 & 0.450 & 0.480 & 0.510 & 0.500 & 0.500 & 0.540 \\
NATOPS                    & 0.900 & 0.911 & 0.933 & 0.944 & 0.894 & 0.756 & 0.733 \\
PenDigits                 & 0.959 & 0.962 & 0.968 & 0.962 & 0.968 & 0.966 & --    \\
PEMS-SF                   & --    & --    & --    & --    & --    & --    & --    \\
PhonemeSpectra            & 0.223 & 0.216 & 0.223 & 0.305 & 0.310 & 0.107 & 0.107 \\
RacketSports              & 0.796 & 0.842 & 0.868 & 0.941 & 0.895 & 0.809 & 0.704 \\
SelfRegulationSCP1        & 0.850 & 0.795 & 0.816 & 0.853 & 0.805 & 0.792 & 0.819 \\
SelfRegulationSCP2        & 0.533 & 0.489 & 0.544 & 0.472 & 0.528 & 0.483 & 0.556 \\
SpokenArabicDigits        & 0.980 & 0.987 & 0.982 & 0.980 & 0.970 & 0.959 & 0.939 \\
StandWalkJump             & 0.400 & 0.400 & 0.400 & 0.333 & 0.400 & 0.400 & 0.333 \\
UWaveGestureLibrary       & 0.906 & 0.909 & 0.906 & 0.891 & 0.797 & 0.853 & 0.897 \\
\midrule
Mean                      & 0.701 & 0.696 & 0.707 & 0.716 & 0.723 & 0.657 & 0.626 \\
\bottomrule
\end{tabular}
\end{table}

\begin{table}[htbp]
\centering
\caption{Per-dataset top-one accuracy of pretrained foundation models using logistic regression.}
\label{tab:pretrained_logreg}
\vspace{0.5em}
\scriptsize
\begin{tabular}{l c c c c c c c}
\toprule
\textbf{Dataset} & \textbf{MOMENT-S} & \textbf{MOMENT-B} & \textbf{MOMENT-L} & \textbf{Mantis} & \textbf{MantisV2} & \textbf{UniTS} & \textbf{NuTime} \\
\midrule
Articulary WordRecognition & 0.993 & 0.990 & 0.983 & 0.993 & 0.993 & 0.910 & 0.980 \\
Atrial Fibrillation       & 0.133 & 0.133 & 0.267 & 0.200 & 0.133 & 0.467 & 0.267 \\
BasicMotions              & 1.000 & 1.000 & 0.975 & 1.000 & 1.000 & 0.925 & 0.750 \\
Character Trajectories    & 0.980 & 0.976 & 0.977 & 0.974 & 0.968 & 0.950 & 0.950 \\
Cricket                   & 0.986 & 0.986 & 0.986 & 0.986 & 0.986 & 0.861 & 0.986 \\
DuckDuckGeese             & 0.480 & 0.500 & 0.540 & 0.420 & 0.500 & --    & 0.240 \\
Eigen Worms               & 0.763 & 0.718 & 0.702 & 0.801 & 0.809 & 0.649 & 0.710 \\
Epilepsy                  & 1.000 & 1.000 & 0.986 & 0.993 & 1.000 & 0.935 & 0.884 \\
ERing                     & 0.941 & 0.915 & 0.900 & 0.963 & 0.978 & 0.852 & 0.867 \\
EthanolConcentration      & 0.331 & 0.350 & 0.338 & 0.304 & 0.369 & 0.262 & 0.601 \\
Face Detection            & --    & --    & --    & --    & --    & 0.580 & --    \\
FingerMovements           & 0.650 & 0.570 & 0.570 & 0.600 & 0.610 & 0.490 & 0.540 \\
HandMovementDirection     & 0.338 & 0.324 & 0.365 & 0.203 & 0.216 & 0.419 & 0.365 \\
Handwriting               & 0.287 & 0.317 & 0.272 & 0.472 & 0.361 & 0.204 & 0.148 \\
Heartbeat                 & 0.732 & 0.727 & 0.717 & 0.805 & 0.795 & 0.644 & 0.702 \\
Japanese Vowels           & 0.960 & 0.954 & 0.949 & 0.976 & 0.970 & 0.900 & 0.792 \\
Libras                    & 0.872 & 0.861 & 0.872 & 0.889 & 0.911 & 0.728 & 0.761 \\
LSST                      & 0.588 & 0.579 & 0.605 & 0.654 & 0.665 & 0.469 & 0.412 \\
MotorImagery              & 0.500 & 0.420 & 0.480 & 0.550 & 0.490 & 0.480 & 0.530 \\
NATOPS                    & 0.922 & 0.922 & 0.933 & 0.928 & 0.867 & 0.717 & 0.667 \\
PenDigits                 & 0.966 & 0.971 & 0.970 & 0.966 & 0.969 & 0.965 & --    \\
PEMS-SF                   & --    & --    & --    & --    & --    & --    & --    \\
PhonemeSpectra            & 0.219 & 0.198 & 0.218 & 0.307 & 0.327 & 0.100 & 0.116 \\
RacketSports              & 0.836 & 0.855 & 0.770 & 0.934 & 0.921 & 0.770 & 0.711 \\
SelfRegulationSCP1        & 0.843 & 0.805 & 0.812 & 0.823 & 0.805 & 0.812 & 0.802 \\
SelfRegulationSCP2        & 0.483 & 0.511 & 0.544 & 0.511 & 0.511 & 0.494 & 0.528 \\
SpokenArabicDigits        & 0.965 & 0.976 & 0.963 & 0.972 & 0.926 & 0.945 & 0.902 \\
StandWalkJump             & --    & --    & --    & --    & --    & 0.333 & 0.333 \\
UWaveGestureLibrary       & 0.878 & 0.881 & 0.878 & 0.903 & 0.794 & 0.828 & 0.884 \\
\midrule
Mean                      & 0.717 & 0.709 & 0.714 & 0.736 & 0.726 & 0.655 & 0.632 \\
\bottomrule
\end{tabular}
\end{table}

\begin{table}[htbp]
\centering
\caption{Per-dataset top-one accuracy of pretrained foundation models using RBF SVM.}
\label{tab:pretrained_svm}
\vspace{0.5em}
\scriptsize
\begin{tabular}{l c c c c c c c}
\toprule
\textbf{Dataset} & \textbf{MOMENT-S} & \textbf{MOMENT-B} & \textbf{MOMENT-L} & \textbf{Mantis} & \textbf{MantisV2} & \textbf{UniTS} & \textbf{NuTime} \\
\midrule
Articulary WordRecognition & 0.990 & 0.983 & 0.987 & 0.993 & 0.993 & 0.903 & 0.987 \\
Atrial Fibrillation       & 0.067 & 0.200 & 0.133 & 0.267 & 0.133 & 0.333 & 0.400 \\
BasicMotions              & 1.000 & 1.000 & 1.000 & 1.000 & 1.000 & 0.825 & 0.700 \\
Character Trajectories    & 0.976 & 0.977 & 0.973 & 0.969 & 0.965 & 0.948 & 0.949 \\
Cricket                   & 0.958 & 0.972 & 0.958 & 1.000 & 0.986 & 0.833 & 0.972 \\
DuckDuckGeese             & 0.380 & 0.380 & 0.400 & 0.440 & 0.420 & --    & 0.180 \\
Eigen Worms               & 0.702 & 0.710 & 0.656 & 0.733 & 0.794 & 0.557 & 0.710 \\
Epilepsy                  & 1.000 & 0.993 & 0.993 & 1.000 & 1.000 & 0.957 & 0.920 \\
ERing                     & 0.937 & 0.900 & 0.841 & 0.948 & 0.970 & 0.878 & 0.856 \\
EthanolConcentration      & 0.285 & 0.228 & 0.281 & 0.312 & 0.392 & 0.278 & 0.407 \\
Face Detection            & --    & --    & --    & --    & --    & 0.564 & --    \\
FingerMovements           & 0.560 & 0.520 & 0.550 & 0.550 & 0.560 & 0.490 & 0.560 \\
HandMovementDirection     & 0.230 & 0.243 & 0.257 & 0.270 & 0.257 & 0.419 & 0.257 \\
Handwriting               & 0.174 & 0.200 & 0.159 & 0.281 & 0.229 & 0.215 & 0.152 \\
Heartbeat                 & 0.722 & 0.722 & 0.722 & 0.810 & 0.795 & 0.688 & 0.722 \\
Japanese Vowels           & 0.949 & 0.951 & 0.938 & 0.973 & 0.965 & 0.862 & 0.795 \\
Libras                    & 0.833 & 0.844 & 0.844 & 0.867 & 0.894 & 0.728 & 0.811 \\
LSST                      & 0.613 & 0.616 & 0.623 & 0.669 & 0.676 & 0.472 & 0.440 \\
MotorImagery              & 0.500 & 0.500 & 0.500 & 0.570 & 0.500 & 0.500 & 0.560 \\
NATOPS                    & 0.894 & 0.839 & 0.872 & 0.933 & 0.894 & 0.711 & 0.689 \\
PenDigits                 & 0.945 & 0.959 & 0.969 & 0.966 & 0.963 & 0.972 & --    \\
PEMS-SF                   & --    & --    & --    & --    & --    & --    & --    \\
PhonemeSpectra            & 0.208 & 0.214 & 0.219 & 0.298 & 0.290 & 0.112 & 0.112 \\
RacketSports              & 0.783 & 0.829 & 0.803 & 0.934 & 0.934 & 0.750 & 0.770 \\
SelfRegulationSCP1        & 0.819 & 0.816 & 0.816 & 0.819 & 0.799 & 0.788 & 0.833 \\
SelfRegulationSCP2        & 0.467 & 0.494 & 0.550 & 0.533 & 0.511 & 0.494 & 0.522 \\
SpokenArabicDigits        & 0.955 & 0.966 & 0.945 & 0.974 & 0.899 & 0.956 & 0.917 \\
StandWalkJump             & 0.400 & 0.467 & 0.533 & 0.333 & 0.400 & 0.400 & 0.333 \\
UWaveGestureLibrary       & 0.875 & 0.853 & 0.922 & 0.866 & 0.791 & 0.825 & 0.900 \\
\midrule
Mean                      & 0.675 & 0.681 & 0.683 & 0.715 & 0.704 & 0.647 & 0.633 \\
\bottomrule
\end{tabular}
\end{table}

\begin{table}[htbp]
\centering
\caption{Per-dataset top-one accuracy of pretrained foundation models using random forest.}
\label{tab:pretrained_rf}
\vspace{0.5em}
\scriptsize
\begin{tabular}{l c c c c c c c}
\toprule
\textbf{Dataset} & \textbf{MOMENT-S} & \textbf{MOMENT-B} & \textbf{MOMENT-L} & \textbf{Mantis} & \textbf{MantisV2} & \textbf{UniTS} & \textbf{NuTime} \\
\midrule
Articulary WordRecognition & 0.990 & 0.993 & 0.993 & 0.993 & 0.993 & 0.867 & 0.973 \\
Atrial Fibrillation       & 0.133 & 0.133 & 0.133 & 0.200 & 0.067 & 0.333 & 0.267 \\
BasicMotions              & 1.000 & 1.000 & 0.975 & 1.000 & 1.000 & 0.900 & 0.725 \\
Character Trajectories    & 0.972 & 0.965 & 0.966 & 0.953 & 0.942 & 0.916 & 0.901 \\
Cricket                   & 0.986 & 1.000 & 0.986 & 1.000 & 0.986 & 0.792 & 0.944 \\
DuckDuckGeese             & 0.480 & 0.360 & 0.380 & 0.420 & 0.500 & --    & 0.180 \\
Eigen Worms               & 0.763 & 0.695 & 0.679 & 0.725 & 0.786 & 0.565 & 0.672 \\
Epilepsy                  & 1.000 & 0.993 & 0.993 & 0.993 & 1.000 & 0.957 & 0.833 \\
ERing                     & 0.933 & 0.878 & 0.870 & 0.956 & 0.952 & 0.878 & 0.874 \\
EthanolConcentration      & 0.354 & 0.308 & 0.331 & 0.259 & 0.395 & 0.278 & 0.262 \\
Face Detection            & --    & --    & --    & --    & --    & 0.537 & --    \\
FingerMovements           & 0.590 & 0.610 & 0.530 & 0.580 & 0.530 & 0.550 & 0.550 \\
HandMovementDirection     & 0.216 & 0.162 & 0.243 & 0.297 & 0.243 & 0.392 & 0.311 \\
Handwriting               & 0.208 & 0.235 & 0.201 & 0.328 & 0.289 & 0.169 & 0.115 \\
Heartbeat                 & 0.737 & 0.727 & 0.742 & 0.785 & 0.815 & 0.717 & 0.717 \\
Japanese Vowels           & 0.927 & 0.935 & 0.916 & 0.962 & 0.951 & 0.811 & 0.765 \\
Libras                    & 0.883 & 0.883 & 0.861 & 0.883 & 0.872 & 0.683 & 0.778 \\
LSST                      & 0.559 & 0.580 & 0.590 & 0.608 & 0.633 & 0.446 & 0.393 \\
MotorImagery              & 0.520 & 0.470 & 0.490 & 0.530 & 0.560 & 0.540 & 0.510 \\
NATOPS                    & 0.878 & 0.878 & 0.883 & 0.922 & 0.878 & 0.633 & 0.644 \\
PenDigits                 & 0.942 & 0.948 & 0.962 & 0.940 & 0.943 & 0.957 & --    \\
PEMS-SF                   & --    & --    & --    & --    & --    & --    & --    \\
PhonemeSpectra            & 0.198 & 0.205 & 0.195 & 0.275 & 0.281 & 0.101 & 0.100 \\
RacketSports              & 0.763 & 0.836 & 0.789 & 0.914 & 0.882 & 0.651 & 0.711 \\
SelfRegulationSCP1        & 0.826 & 0.823 & 0.826 & 0.802 & 0.778 & 0.816 & 0.833 \\
SelfRegulationSCP2        & 0.472 & 0.456 & 0.511 & 0.500 & 0.533 & 0.533 & 0.528 \\
SpokenArabicDigits        & 0.919 & 0.915 & 0.856 & 0.930 & 0.795 & 0.913 & 0.819 \\
StandWalkJump             & 0.467 & 0.400 & 0.600 & 0.400 & 0.333 & 0.667 & 0.600 \\
UWaveGestureLibrary       & 0.866 & 0.853 & 0.912 & 0.856 & 0.738 & 0.791 & 0.884 \\
\midrule
Mean                      & 0.688 & 0.676 & 0.682 & 0.704 & 0.692 & 0.644 & 0.611 \\
\bottomrule
\end{tabular}
\end{table}

\newpage

\section{Time-series-specific baseline results}
\label{sec:ts_results}

Table~\ref{tab:ts_specific} reports classification accuracy for ROCKET, Quant, and Hydra on all 30 UEA datasets. These methods provide external time-series-specific reference points and are not architecture-matched controls for the foundation models.

\begin{table}[htbp]
\centering
\caption{Per-dataset classification accuracy of the time-series-specific reference methods on all 30 UEA datasets.}
\label{tab:ts_specific}
\vspace{0.5em}
\small
\begin{tabular}{l c c c}
\toprule
\textbf{Dataset} & \textbf{ROCKET} & \textbf{Quant} & \textbf{Hydra} \\
\midrule
Articulary WordRecognition & 0.9933 & 0.9800 & 0.9933 \\
Atrial Fibrillation       & 0.0667 & 0.2000 & 0.1333 \\
BasicMotions              & 1.0000 & 1.0000 & 1.0000 \\
Character Trajectories    & 0.9923 & 0.9833 & 0.9861 \\
Cricket                   & 1.0000 & 0.9722 & 0.9861 \\
DuckDuckGeese             & 0.5400 & 0.4600 & 0.7000 \\
Eigen Worms               & 0.9008 & 0.8931 & 0.9695 \\
Epilepsy                  & 0.9855 & 1.0000 & 1.0000 \\
ERing                     & 0.9815 & 0.9778 & 0.9852 \\
EthanolConcentration      & 0.4183 & 0.6882 & 0.5209 \\
Face Detection            & 0.5965 & 0.6285 & 0.5287 \\
FingerMovements           & 0.5700 & 0.5700 & 0.5300 \\
HandMovementDirection     & 0.5270 & 0.4730 & 0.2432 \\
Handwriting               & 0.5847 & 0.3106 & 0.4729 \\
Heartbeat                 & 0.7756 & 0.7512 & 0.7610 \\
Insect Wingbeat           & 0.4016 & 0.7272 & 0.6780 \\
Japanese Vowels           & 0.8595 & 0.9622 & 0.9838 \\
Libras                    & 0.9056 & 0.8889 & 0.9389 \\
LSST                      & 0.6444 & 0.6464 & 0.6123 \\
MotorImagery              & 0.5100 & 0.5200 & 0.5000 \\
NATOPS                    & 0.8778 & 0.8722 & 0.8556 \\
PEMS-SF                   & 0.8497 & 0.9884 & 0.8150 \\
PenDigits                 & 0.9840 & 0.9740 & 0.9605 \\
PhonemeSpectra            & 0.2708 & 0.2726 & 0.3269 \\
Racket Sports             & 0.9079 & 0.8882 & 0.8947 \\
SelfRegulationSCP1        & 0.8532 & 0.8225 & 0.8703 \\
SelfRegulationSCP2        & 0.5389 & 0.5056 & 0.5889 \\
Spoken Arabic Digits      & 0.9984 & 0.9818 & 0.9891 \\
StandWalk Jump            & 0.4667 & 0.2667 & 0.3333 \\
UWaveGestureLibrary       & 0.9315 & 0.8875 & 0.9000 \\
\midrule
Mean                      & 0.7311 & 0.7364 & 0.7353 \\
\bottomrule
\end{tabular}
\end{table}

\section{Controlled CPU diagnostic}
\label{app:cpu_diagnostic}

We report a controlled single-thread CPU diagnostic using synthetic
standard-normal \texttt{float32} input of shape
$[B,C,T]=[1,3,512]$ and batch size one. Measurements are performed on an
Intel Core Ultra 7 265H CPU using Windows~11 Pro, Python~3.11.16, and
PyTorch~2.5.1+\texttt{cpu}. Since pretrained and randomly initialized
variants share the same architecture and inference path, each architecture
is measured only once.

For MOMENT and Mantis, latency covers the frozen encoder and its aggregation
operation. For ROCKET, Hydra, and Quant, latency covers the already-fitted
feature transformation, including scaling where applicable. All measurements
exclude model loading, input generation, disk I/O, and final classifier
prediction. The \texttt{aeon} methods use \texttt{aeon}~1.3.0 and
scikit-learn~1.7.2; their fit time is reported separately on 100 balanced
synthetic training cases.

Unless otherwise stated, we use 20 warm-up calls followed by 100 timed calls. NuTime and UniTS are omitted because their released multivariate inference paths are not directly compatible with the common $[1,3,512]$ input protocol. The measurements below cover different pipeline components and should therefore be interpreted as implementation-level diagnostics rather than an end-to-end cross-method ranking.

\begin{table}[htbp]
\caption{Controlled single-thread CPU diagnostic on synthetic
\texttt{float32} $[1,3,512]$ input. $n$ denotes the number of timed calls.
The \texttt{aeon} fit time is measured separately on 100 synthetic training
cases and is not comparable to TSFM pretraining.}
\label{tab:cpu_diagnostic}
\centering
\small
\begin{tabular}{l r r r}
\toprule
Method & Median latency (ms) & $n$ & \texttt{aeon} fit (s)$^\dagger$ \\
\midrule
\multicolumn{4}{l}{\emph{Frozen TSFM encoder and aggregation}} \\
MantisV2 & 15.88 & 100 & -- \\
Mantis & 18.61 & 100 & -- \\
MOMENT-Small & 82.48 & 100 & -- \\
MOMENT-Base$^\ddagger$ & 354.91 & 50 & -- \\
MOMENT-Large$^\ast$ & 1354.18 & 5 & -- \\
\midrule
\multicolumn{4}{l}{\emph{Already-fitted \texttt{aeon} feature transformation}} \\
Hydra (transform + sparse scale) & 14.26 & 100 & 1.64 \\
Quant (transform) & 28.99 & 100 & 1.75 \\
ROCKET (transform + scale) & 46.83 & 100 & 4.48 \\
\bottomrule
\end{tabular}

\vspace{1mm}
\footnotesize
$^\dagger$Separate fit phase on 100 synthetic training cases.
$^\ddagger$10 warm-up calls.
$^\ast$Two warm-up calls and five timed calls.
\end{table}

For the compatible neural models, we additionally count registered
matrix-multiplication and convolution operations using
\texttt{torch.utils.flop\_counter.FlopCounterMode}, with each
multiply--accumulate counted as two operations. The counter does not include
normalization, softmax, activations, interpolation, indexing, allocation, or
non-Torch operations. The reported values are therefore partial operation
counts rather than total FLOPs. We do not report corresponding values for
ROCKET, Hydra, or Quant because substantial parts of their implementations
are not covered by the counter.

\begin{table}[htbp]
\caption{Registered matrix-multiplication and convolution operations for one
frozen neural-encoder forward pass on synthetic $[1,3,512]$ input. Values are
partial instrumented counts rather than total FLOPs.}
\label{tab:cpu_partial_flops}
\centering
\small
\begin{tabular}{l r}
\toprule
Neural encoder & Registered operations (GFLOPs) \\
\midrule
MantisV2 & 0.482 \\
Mantis & 0.874 \\
MOMENT-Small & 7.400 \\
MOMENT-Base & 33.070 \\
MOMENT-Large & 119.591 \\
\bottomrule
\end{tabular}
\end{table}

Within the MOMENT family, median latency increases from 82.48\,ms for Small
to 354.91\,ms for Base and 1354.18\,ms for Large, together with an increase
in registered operations from 7.400 to 119.591\,GFLOPs. MantisV2 and Mantis
record substantially lower encoder-forward latencies of 15.88 and 18.61\,ms,
respectively. Under the separate fitted-\texttt{aeon}-transform scope, Hydra,
Quant, and ROCKET record median latencies of 14.26, 28.99, and 46.83\,ms.

Because the two groups measure different pipeline components, these values should not be used as a direct end-to-end efficiency ranking. They provide controlled context for the specified synthetic input and should not be extrapolated to native UEA sequence lengths, channel counts, batching strategies, hardware platforms, or complete deployment pipelines. These diagnostics measure latency and partial operation counts only; memory usage is not measured, so no conclusions are drawn about memory efficiency across methods.

\end{document}